\documentclass{article} 
\usepackage{nips14submit_e,times}
\usepackage{hyperref}
\usepackage{url}
\usepackage[numbers]{natbib}
\usepackage[pdftex]{graphicx}
\usepackage{amsmath}
\usepackage{amssymb}
\usepackage{wrapfig}

\title{Microscopic Advances with Large-Scale Learning: Stochastic Optimization for Cryo-EM}

\author{
Ali Punjani \\
Department of Computer Science \\
University of Toronto \\
\texttt{alipunjani@cs.toronto.edu} \\
\And 
Marcus A.~Brubaker \\
Department of Computer Science \\
University of Toronto \\
\texttt{mbrubake@cs.toronto.edu}
}

\nipsfinalcopy 

\begin{document}

\newcommand{\fix}[1]{\marginpar{\footnotesize FIX: #1}}
\newcommand{\new}{\marginpar{NEW}}

\newcommand{\marcus}[1]{\textcolor{blue}{\textbf{[Marcus: #1]}}}
\newcommand{\ali}[1]{\textcolor{green}{\textbf{[Ali: #1]}}}
\newcommand{\comment}[1]{}

\newcommand{\eg}{{\emph{e.g.}}}
\newcommand{\ie}{{\emph{i.e.}}}

\newcommand{\R}{\mathbb{R}}
\newcommand{\SO}{\mathcal{SO}(3)}
\newcommand{\eye}{\mathbf{I}}

\newcommand{\projdir}{\mathbf{R}}
\newcommand{\proj}[1]{ \mathbf{P}_{ { #1 } } }
\newcommand{\ftproj}[1]{ \tilde{\mathbf{P}}_{ { #1 } } }

\newcommand{\ctfparam}{\theta}
\newcommand{\ctf}[1]{\mathbf{C}_{#1}}
\newcommand{\ftctf}[1]{\tilde{\mathbf{C}}_{#1}}

\newcommand{\shiftdir}{\mathbf{t}}
\newcommand{\shift}[1]{ \mathbf{S}_{ { #1 } } }
\newcommand{\ftshift}[1]{ \tilde{\mathbf{S}}_{ { #1 } } }

\newcommand{\img}{\mathcal{I}}
\newcommand{\ftimg}{\tilde{\mathcal{I}}}
\newcommand{\ftdensity}{\tilde{\mathcal{V}}}
\newcommand{\density}{\mathcal{V}}

\newcommand{\noiseStd}{\sigma}

\newcommand{\data}{\mathfrak{D}}

\maketitle

\begin{abstract}
Determining the 3D structures of biological molecules is a key problem for 
both biology and medicine. 
Electron Cryomicroscopy (Cryo-EM) is a promising technique for structure 
estimation which relies heavily on computational methods to reconstruct
3D structures from 2D images.
This paper introduces the challenging Cryo-EM density estimation problem 
as a novel application for stochastic optimization techniques.
Structure discovery is formulated as MAP estimation in a
probabilistic latent-variable model, resulting
in an optimization problem to which an array of seven stochastic optimization 
methods are applied.
The methods are tested on both real and synthetic data, with some methods
recovering reasonable structures in less than one epoch from a random
initialization.
Complex quasi-Newton methods are found to converge more slowly than simple
gradient-based methods,  but all stochastic methods are found to converge to
similar optima. 
This method represents a major improvement over existing methods as it is significantly
faster and is able to converge from a random initialization.
\end{abstract}

\section{Introduction}
Discovering the 3D structure of molecules such as proteins and viruses is an
important problem in biology and medicine.
Biological macromolecules are composed of chains of simpler monomers, and the 
conformation or ``folding'' of these chains into a 3D structure determines its
specific function and properties.
Traditional approaches to estimating 3D structures, such as X-ray crystallography
or nuclear magnetic resonance (NMR) spectroscopy, have fundamental limitations.
X-ray crystallography requires a crystal of the target molecule; these are 
difficult to grow at best, and often impossible \citep{Rupp2009}.
NMR doesn't require a special form of the target, but is limited to relatively
small molecules, preventing the study of important biological complexes
\citep{Keeler2010}.
Electron Cryomicroscopy (Cryo-EM) is an emerging experimental methodology for
structure determination which is able to measure medium to large-sized
molecules in a native state, \ie, without a need for crystallization or
non-native solvents \citep{Chiu1993}.
However, Cryo-EM raises challenging computational problems, one of which 
we attempt to address here.

\begin{figure}
\centering
\includegraphics[width=\textwidth]{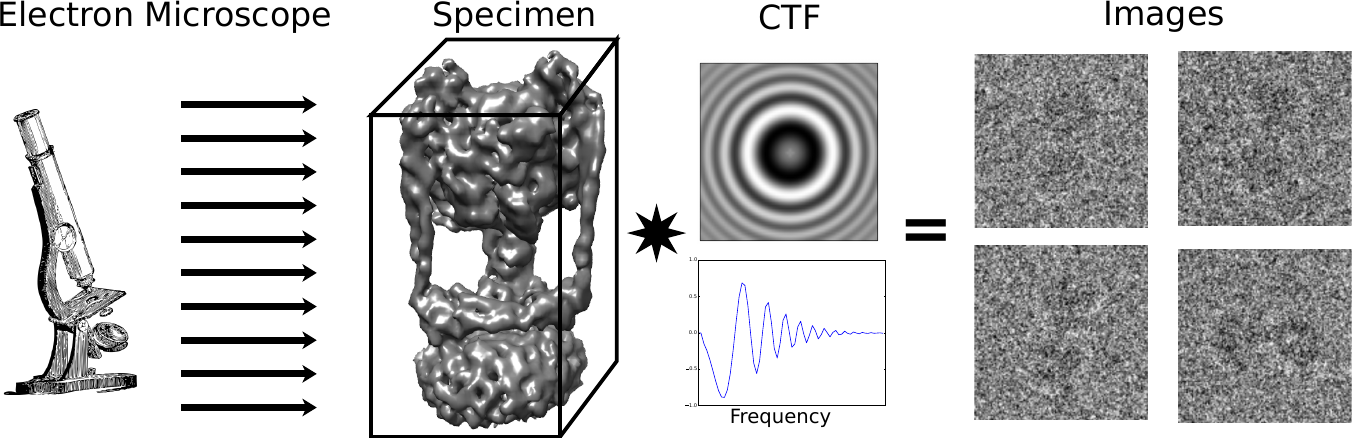}
\caption{
\label{fig:cryoimgs}
A generative image formation model in Cryo-EM.  
The electron beam results in an orthographic integral projection of the
electron density of the specimen.  This projection is modulated by the
Contrast Transfer Function (CTF) and corrupted with noise.
The images pictured here are from a dataset of real
particle images of protein \emph{ATP synthase}, showcasing the low SNR
typical in Cryo-EM.
The zeros in the CTF make estimation
particularly challenging however their location will vary
as a function of experimental parameters.
Particle images and density from \citep{Lau2012}.
}
\end{figure}

In Cryo-EM, a purified solution of target molecules is frozen in a thin film
and imaged under a transmission electron microscope.
The scattering of the electrons as they pass through the sample is measured, 
producing images in which individual molecules are visible.
Each particle image is related to an orthographic, integral projection of the
electron density of the target molecule, but the direction of projection is unknown.
The captured image is further corrupted by 
destructive interference in the electron microscope imaging process
and, due to the sensitive nature of biological specimens, the radiation dosage
is kept to a minimum, leading to particle images with extremely low
signal-to-noise ratios (SNR), typically around 0.05 \citep{Baxter2009}.
The Cryo-EM imaging method, including typical particle images, is illustrated
in Figure \ref{fig:cryoimgs}.

The computational task in Cryo-EM is to estimate the 3D electron density
of a target molecule, given a set of particle images.
This is similar to density estimation in Computed Tomography (CT), however
in CT the projection direction of each image is known \citep{Hsieh2003}.
Together, unknown orientations and image corruption make Cryo-EM density
estimation a challenging  problem.
Inspecting the real particle images in Figure \ref{fig:cryoimgs} makes 
this clear -- to the human eye the coarse dumbbell shape of the molecule is
barely visible but finer details, like the presence of three stalks, are
practically imperceptable.

In this paper we explore the use of stochastic optimization techniques for
Cryo-EM density estimation.
To do this we introduce a probabilistic latent-variable model of image formation
in Cryo-EM in which we seek the maximum-a-posteriori (MAP) estimate of 
the electron density.
We then formulate electron density estimation as a stochastic optimization
problem.
We show that this approach leads to significant speed gains, providing the
ability to compute density estimates in just a few hours where existing
approaches would take days or weeks.
Further, our results show that the stochastic optimization approach is dramatically
less  sensitive to initialization than previous methods.
While these previous approaches can be strongly biased by bad initialization
\citep{Henderson2012}, our method is able to quickly converge from a random
initialization.
We explore an array of stochastic optimization algorithms and compare
their performance on a new class of objective functions which arises from the
probabilistic model.

We believe that Cryo-EM is an important and challenging problem which, with
a few exceptions (\eg, \citep{Mallick2006}), has seen little attention in
the machine learning and computer vision communities.
Our mixed results of comparing stochastic optimization methods
also suggest that Cryo-EM may serve as an important benchmark problem for 
new stochastic optimization algorithms.
While stochastic optimization algorithms have had significant success,
their application has typically been limited to a small set of objective
functions.
With this paper we hope to spark interest in this problem at large and in
Cryo-EM density estimation as a real-world benchmark for stochastic
optimization methods.

\section{Background and Related Work} \label{sec:background}

Traditional approaches to Cryo-EM density estimation, \eg, \citep{Grigorieff2007},
aim to solve the problem by iterative refinement.
An initial density estimate is projected in a large number of 
directions and each projection is compared with each particle image.
For each particle, the projection which matches closest is considered the true
orientation of the particle.
Reconstruction of a new density estimate is then based on the Fourier
Slice Theorem (FST) which states that the 2D Fourier transform of a projection
of a density is equal to a slice through the origin of the 3D Fourier
transform of that density, in a plane perpendicular to the projection direction
\citep{Hsieh2003}.
Using the computed orientations of each particle, the new density is estimated by
interpolation and averaging of the observed particle images.
This approach is fundamentally limited as, even under ideal circumstances,
the low SNR of particle images makes accurately identifying the single correct
orientation for each particle nearly impossible resulting in errors in 
the estimated density.
This problem is exacerbated when attempting to refine a density without prior
knowledge of the shape; a poor initialization will result in estimating
a structure which is either clearly wrong (see Figure \ref{fig:baselines}) or,
worse, appears correct but is misleading, resulting in the publication of
incorrectly estimated 3D structures \citep{Henderson2012}.
Finally, and crucially for the case of density estimation with many particle
images, all data is used at each refinement iteration causing these methods
to be slow in general.

Recently Bayesian approaches to density estimation have been
proposed which avoid estimating a single orientation for individual particle
images.
The orientation of each particle is treated as a random variable rather than
simply an unknown parameter, and all orientations are considered by marginalizing 
over these random variables.
The resulting integral is analytically intractable but can be computed
numerically.
Further, the simple interpolation and averaging based off of the FST
is no longer possible and some form of optimization must be performed to
estimate the density. 
Marginalization for Cryo-EM was originally proposed for 2D image alignment by
\cite{Sigworth1998} and 3D reconstruction by \cite{Scheres2007}.
Gradient based optimization with marginalization was originally
proposed by \citet{Jaitly2010} where a batch, gradient-based optimization
was performed but using only a small number of low-noise images which were
found by clustering and averaging individual particle images.
\citet{Scheres2012} used marginalization with a batch Expectation-Maximization
(EM) algorithm in the RELION package with raw particle images.
In this paper we work with a similar generative model as these methods, but show
that with stochastic optimization, significant progress towards a MAP estimate
can be made quickly and with better robustness to initialization.

Stochastic optimization has seen significant theoretical and practical interest
recently.
The fundamental approach of \emph{stochastic gradient descent} (SGD)
remains a popular and often surprisingly effective algorithmic choice.
Momentum based methods improve on SGD by using gradient information from
multiple iterations, allowing for faster traversal of directions with low
curvature \citep{Polyak1964,Nesterov1983,Sutskever2013}.
Natural gradient methods have been developed based on theoretical connections
to the manifold geometry of parameter space through the Fisher information
matrix \citep{Amari1998,Amari2000} and statistical considerations of
accounting for noise in gradient vectors \citep{LeRoux2008,LeRoux2010}.
Higher-order methods have been developed which attempt to use either
approximate \citep{Schraudolph2007,Bordes2009} or analytic Hessians
\citep{Martens2010,LeRoux2010} to speed convergence by explicitly
accounting for curvature.
Most methods have a number of hyper-parameters and are highly sensitive
to their settings; this has motivated a number of attempts to develop
methods which have fewer parameters \citep{Duchi2011,Schaul2013}.
Finally, when operating in the finite-data context, algorithms have been 
developed which can utilize the gradients of all data-points while still
operating on only a limited subset at a time \citep{LeRoux2012,Schmidt2013a}
and have strong convergence results.
A full review of the theoretical results is beyond the scope
of this paper and we refer interested readers to \citep{Bottou2011}.
In this paper we compare a number of these methods and evaluate their
performance and suitability for the given task.
Our results show that while some methods can find good solutions more quickly,
almost all methods converge to a similarly optimal solution, and that simpler 
methods are typically as good or better than more complex and costly ones.

\section{A Probabilistic Model of Cryo-EM}
In order to formulate density estimation as an optimization problem, we
turn to a probabilistic latent variable model with three parts; particle images
are observed variables, their corresponding orientations are unknown latent
variables, and the electron density is an unknown parameter for which we
seek a MAP estimate. 
Each particle image is considered an orthographic, integral projection of electron 
density along the direction of the microscope beam.
This image is corrupted by two phenomenon: the contrast transfer function
(CTF) and noise.

The primary effects of the CTF, resulting from destructive interference, are 
modelled by a modulation in frequency space, \ie, as a linear operator on
the integral projection.
The Fourier spectrum of a typical CTF is shown in Figure
\ref{fig:cryoimgs}.
Note the phase changes and zero crossings which result in missing information
in individual images.
These zero crossings vary with the experimental settings of the microscope, 
and although each sample of target molecules can only be imaged once, different
samples are imaged under different conditions to ensure that every frequency
is captured.
A full review of the CTF is beyond the scope of this paper and we refer
interested readers to \citep{Reimer2008}.
Noise arises due to the low exposure dosages necessary due to sensitive biological
molecules.
This noise is generally well modelled as additive IID Gaussian noise.

To formalize this, we represent an electron density as a 3D grid with density
at each voxel, denoted here as $\density \in \R^{N^3}$ where $N$ is the side length
of the cubic grid.
An integral projection of this density in some orientation $\projdir \in \SO$
can be represented as a linear transformation
$\proj{\projdir} \in \R^{N^2 \times N^3}$.
We assume that each particle image $\img \in \R^{N^2}$ has an associated set
of CTF parameters $\ctfparam$.
As discussed above, the CTF is modelled as a linear operator on the projected
image, denoted here as $\ctf{\ctfparam} \in \R^{N^2 \times N^2}$.
Finally, the particle is subject to a 2D shift $\shiftdir \in \R^2$ in the image
plane as it is not necessarily centered in the image which can, similarly, 
be represented as a linear operator denoted by
$\shift{\shiftdir} \in \R^{N^2 \times N^2}$.
The conditional probability distribution of observing an image is thus
\begin{equation}
p(\img | \ctfparam, \projdir, \shiftdir, \density ) = \mathcal{N}(\img | \shift{\shiftdir} \ctf{\ctfparam} \proj{\projdir} \density, \noiseStd^2 \eye)
\label{eq:BaseLikelihood}
\end{equation}
where $\noiseStd$ is the standard deviation of the noise and
$\mathcal{N}(\mu,\Sigma)$ is the multivariate normal distribution with mean
vector $\mu$ and covariance matrix $\Sigma$.
In practice due to computational considerations Equation 
\ref{eq:BaseLikelihood} is evaluated in Fourier space, making use of the
Fourier Slice Theorem and Parseval's Theorem to obtain
\begin{equation}
p(\ftimg | \ctfparam, \projdir, \shiftdir, \ftdensity ) = \mathcal{N}(\ftimg | \ftshift{\shiftdir} \ftctf{\ctfparam} \ftproj{\projdir} \ftdensity, \noiseStd^2 \eye)
\label{eq:BaseFTLikelihood}
\end{equation}
where $\ftimg$ is the Fourier transform of the image, $\ftshift{\shiftdir}$
is the shift operator in Fourier space (a phase change), $\ftctf{\ctfparam}$ is
the CTF modulation in Fourier space (a diagonal operator), $\ftproj{\projdir}$ is a sinc
interpolation operator which extracts a plane through the origin defined by 
the projection orientation $\projdir$ and $\ftdensity$ is the 3D Fourier
transform of $\density$.
To further speed computation of the likelihood, and because of the level of
noise and the attenuation of high frequencies by the CTF, Equation
\ref{eq:BaseFTLikelihood} is only evaluated using Fourier coefficients up to a
specified maximum frequency.

\begin{figure}
\centering

\hfill
\includegraphics[width=0.13\textwidth]{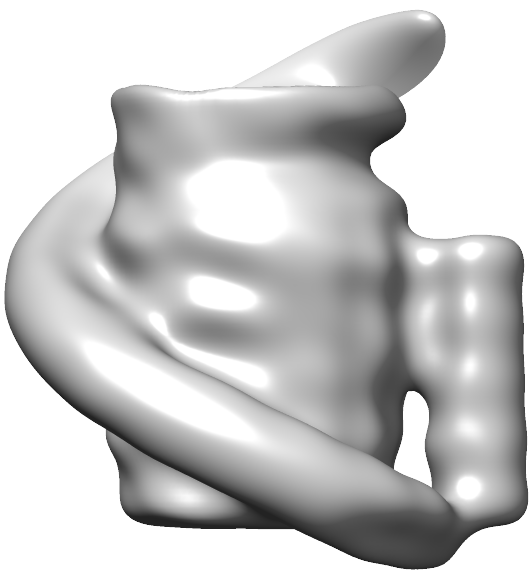}
\hfill
\includegraphics[width=0.145\textwidth]{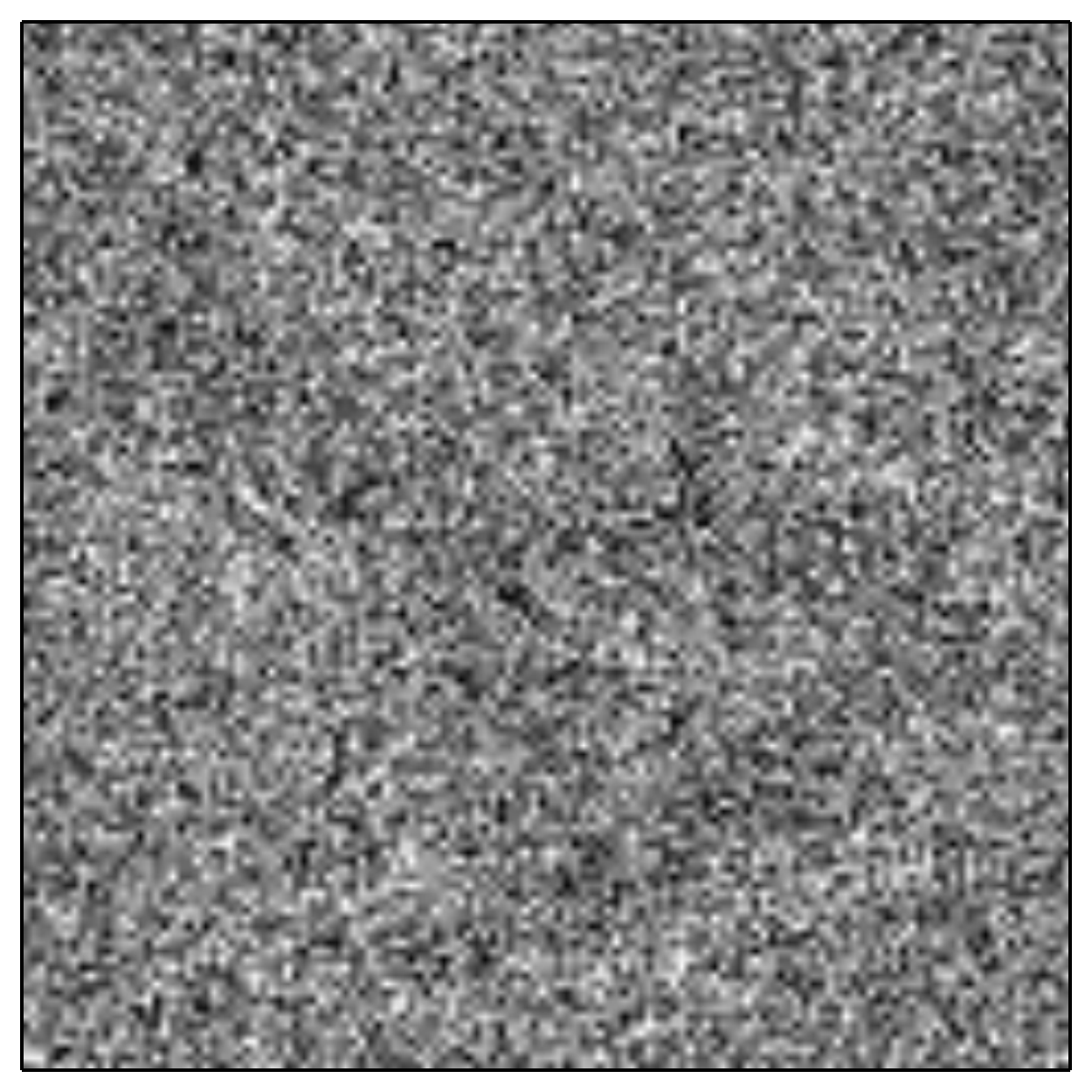}
\includegraphics[width=0.145\textwidth]{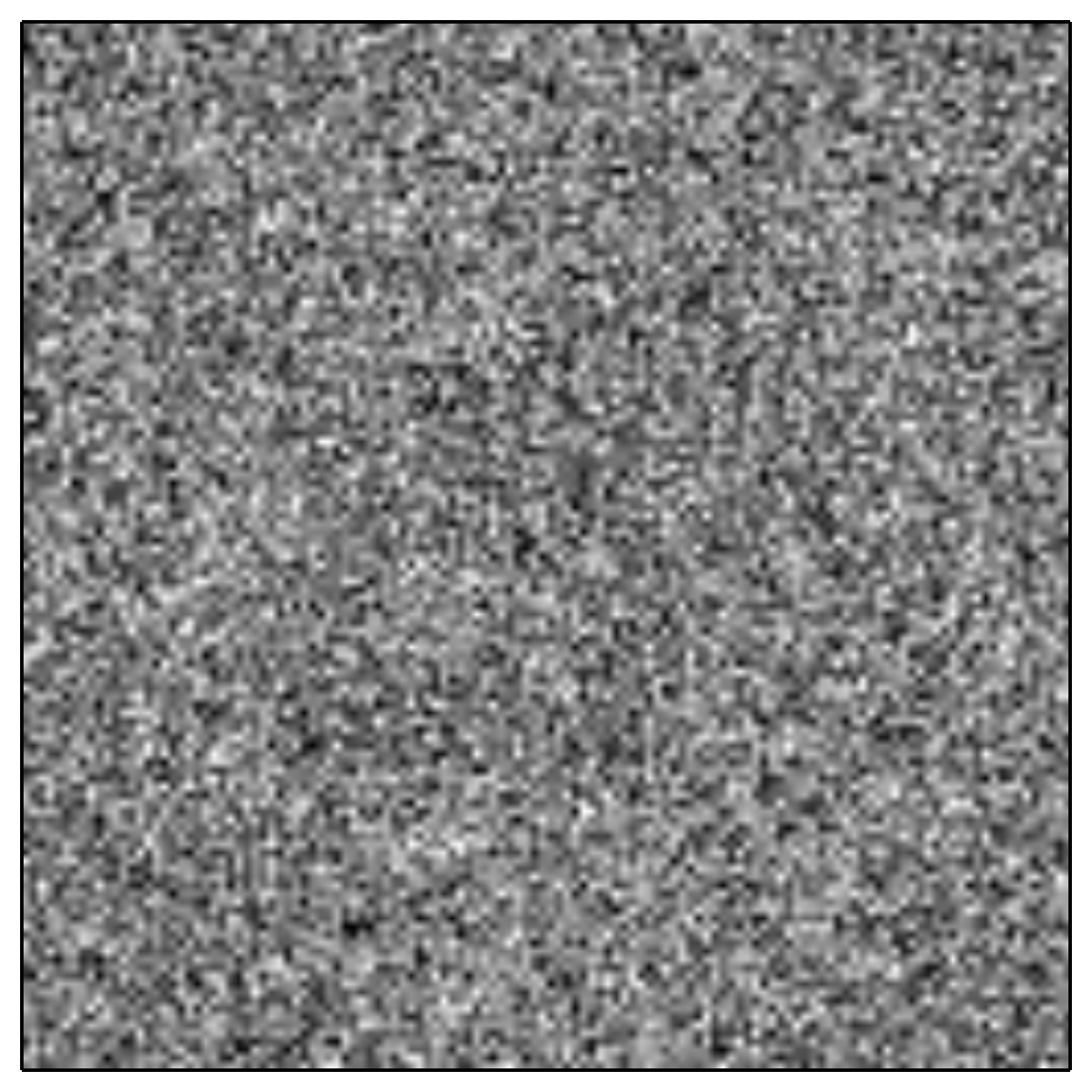}
\includegraphics[width=0.145\textwidth]{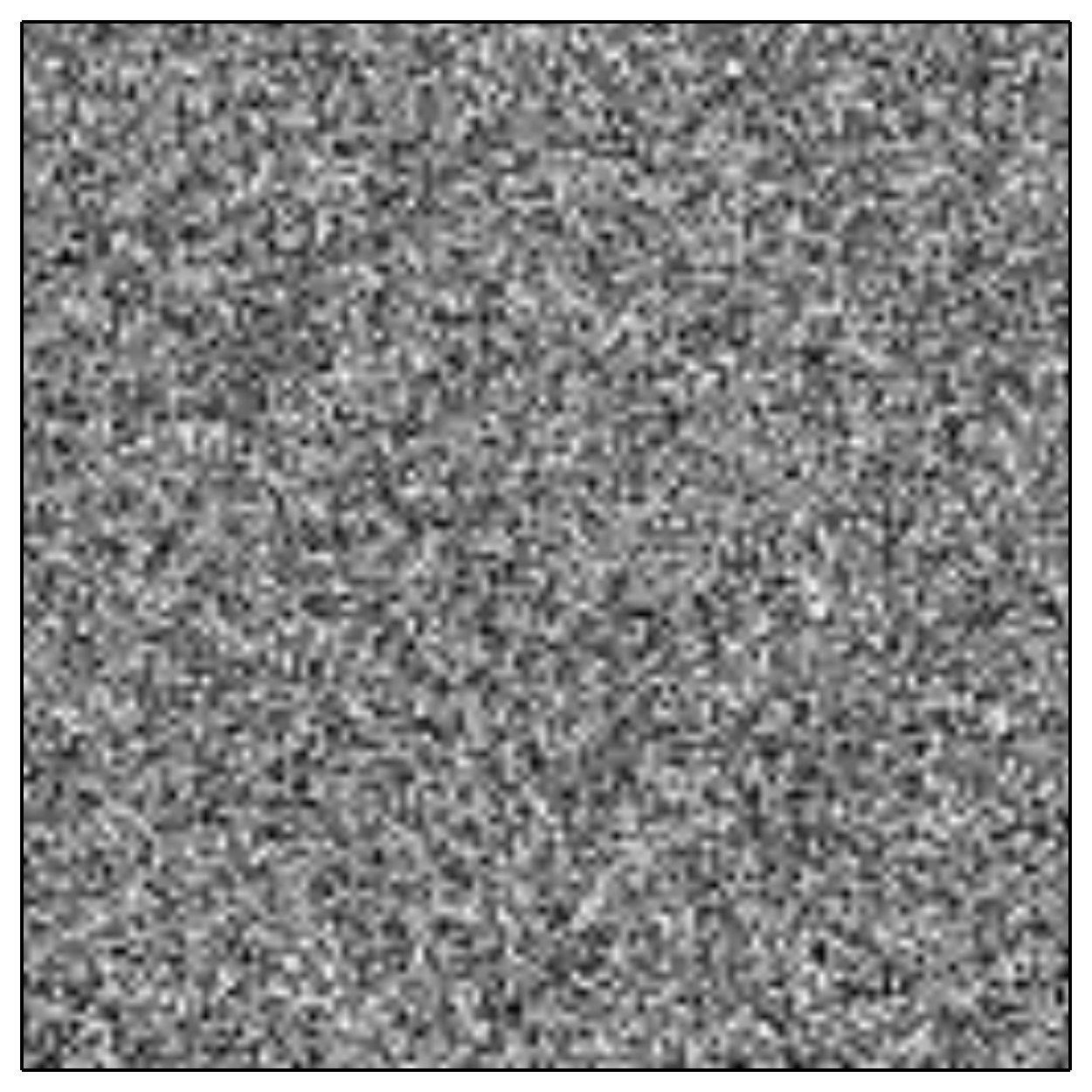}
\includegraphics[width=0.145\textwidth]{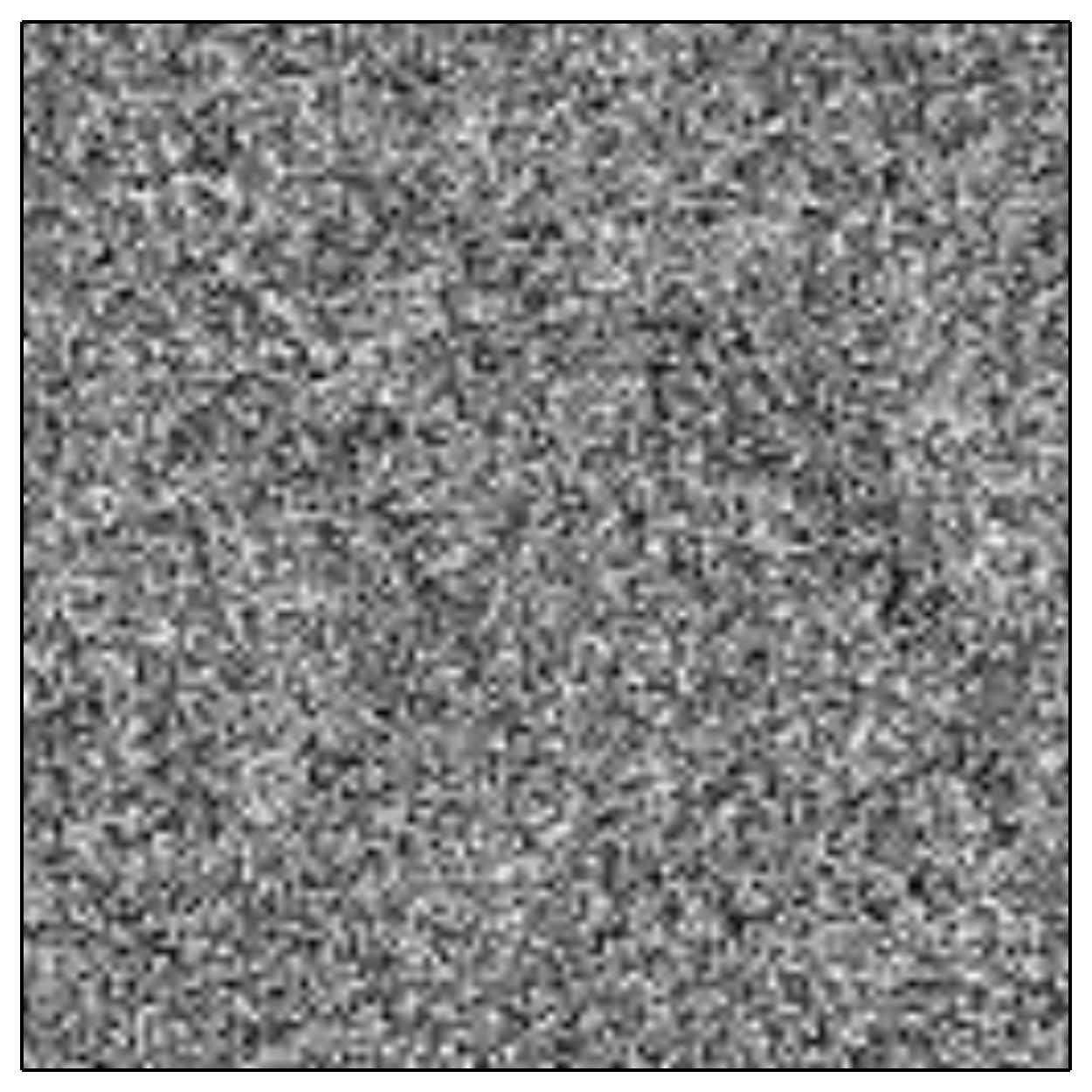}
\includegraphics[width=0.145\textwidth]{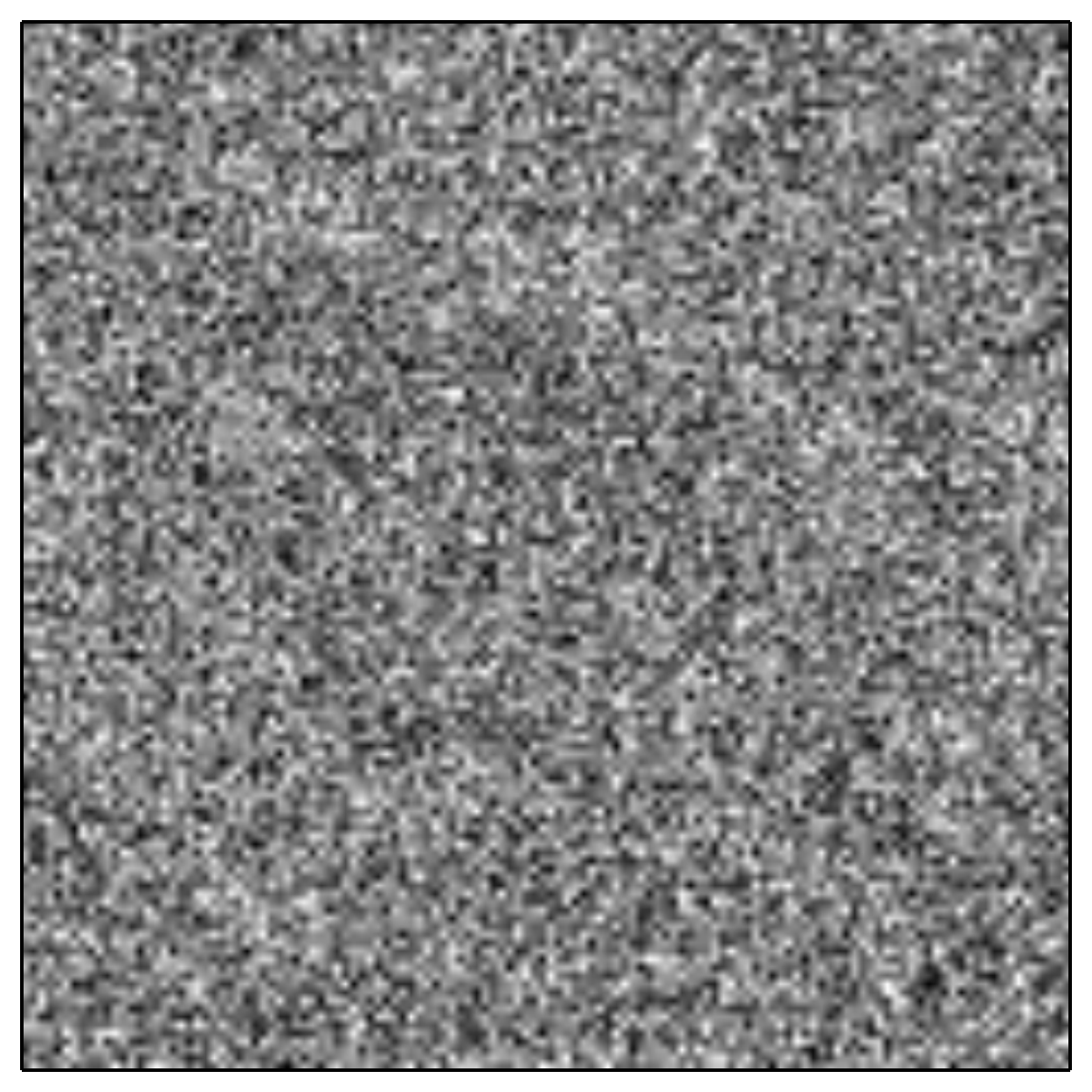}
\hfill
\caption{\label{fig:phantom}
The synthetic density (left) and simulated images
(right) generated by choosing orientations uniformly at random,
applying CTFs selected from a real dataset and adding noise.
}
\vspace{-0.2cm}
\end{figure}

This image formation model provides the conditional probability of observing an 
image $\ftimg$ from a given orientation, shift and density.
The orientation and shift, however, are unknown.
To cope with this, we marginalize over these latent variables
\citep{Jaitly2010,Scheres2012} and, assuming they are independent of each other
and the density $\density$, get
\begin{equation}
p(\ftimg | \ctfparam, \ftdensity) = \int_{\R^2} \int_{\SO} p(\ftimg | \ctfparam, \projdir, \shiftdir, \ftdensity ) p(\projdir) p(\shiftdir) d\projdir d\shiftdir
\end{equation}
where $p(\projdir)$ is the uniform distribution over $\SO$ and we
use a normal distribution truncated based on the size of the particle images
for the shift distribution $p(\shiftdir)$.
This double integral is not analytically tractable and so we resort to
numerical quadrature.
We use Lebedev quadrature over directions of projection in $\mathcal{S}^2$
combined with uniform quadrature over the interval $[0,2\pi)$ to
account for in-plane rotation \citep{Lebedev1999,Graef2009} and uniform
quadrature over the truncated region of $\R^2$ to account for shifts.
Using numerical quadrature, the conditional probability of an 
image is
\begin{equation}
p(\ftimg | \ctfparam, \ftdensity) \approx \sum_{j=1}^{M} w_j p(\ftimg | \ctfparam, \projdir_j, \shiftdir_j, \ftdensity )
\label{eq:ApproxMarginalization}
\end{equation}
where $\{(\projdir_j,\shiftdir_j,w_j)\}_{j=1}^{M}$ are the weighted quadrature
points.
The accuracy of the quadrature scheme, and consequently the value of $M$, is
set automatically based on the specified maximum frequency considered.

Given a set of $K$ images with CTF parameters
$\data = \{(\img_i,\ctfparam_i)\}_{i=1}^{K}$ and assuming conditional
independence of the images, the posterior is
\begin{equation}
p(\density | \data) \propto p(\density) \prod_{i=1}^K p(\ftimg_i | \ctfparam_i, \ftdensity)
\label{eq:Posterior}
\end{equation}
where $p(\density)$ is a prior.
We use a combination of an exponential distribution for positive
density values to encourage sparsity and a generalized normal
distribution to softly penalize negative density values.
Specifically, $p(\density) = \prod_{i=1}^{N^3} p(\density_i)$
where $\density_i$ is the value of the $i$th voxel,
$p(\density_i) \propto e^{-\lambda_+ \density_i}$ if $\density_i \geq 0$, and 
$p(\density_i) \propto e^{-\lambda_- |\density_i|^3}$ if $\density_i < 0$.
Many other choices of prior are possible and is a promising direction for
future research.

Estimating the electron density then corresponds to finding $\density$
which maximizes Equation \ref{eq:Posterior}.
Optimizing this directly is costly due to the marginalization in Equation
\ref{eq:ApproxMarginalization}.
When selecting a maximum frequency cut-off of only 16\% of the Nyquist
limit, this corresponds to approximately $95,000$ quadrature points for the
datasets shown here.
Combined with a dataset size of around $46,000$ particle images, a full
evaluation of the posterior and its gradient takes over a day of computation
time on a modern CPU.
Instead, we formulate this as a stochastic optimization problem.
Taking the log and dropping constant factors the
optimization problem becomes
\begin{equation}
\arg\min_{\density} -\sum_{i=1}^K\left(\log p(\ftimg_i | \ctfparam_i, \ftdensity) + K^{-1}\log p(\density)\right)
\label{eq:OptimProblem}
\end{equation}
which is the standard form for a stochastic optimization problem.

\begin{figure}
\centering

\includegraphics[height=0.198\textwidth]{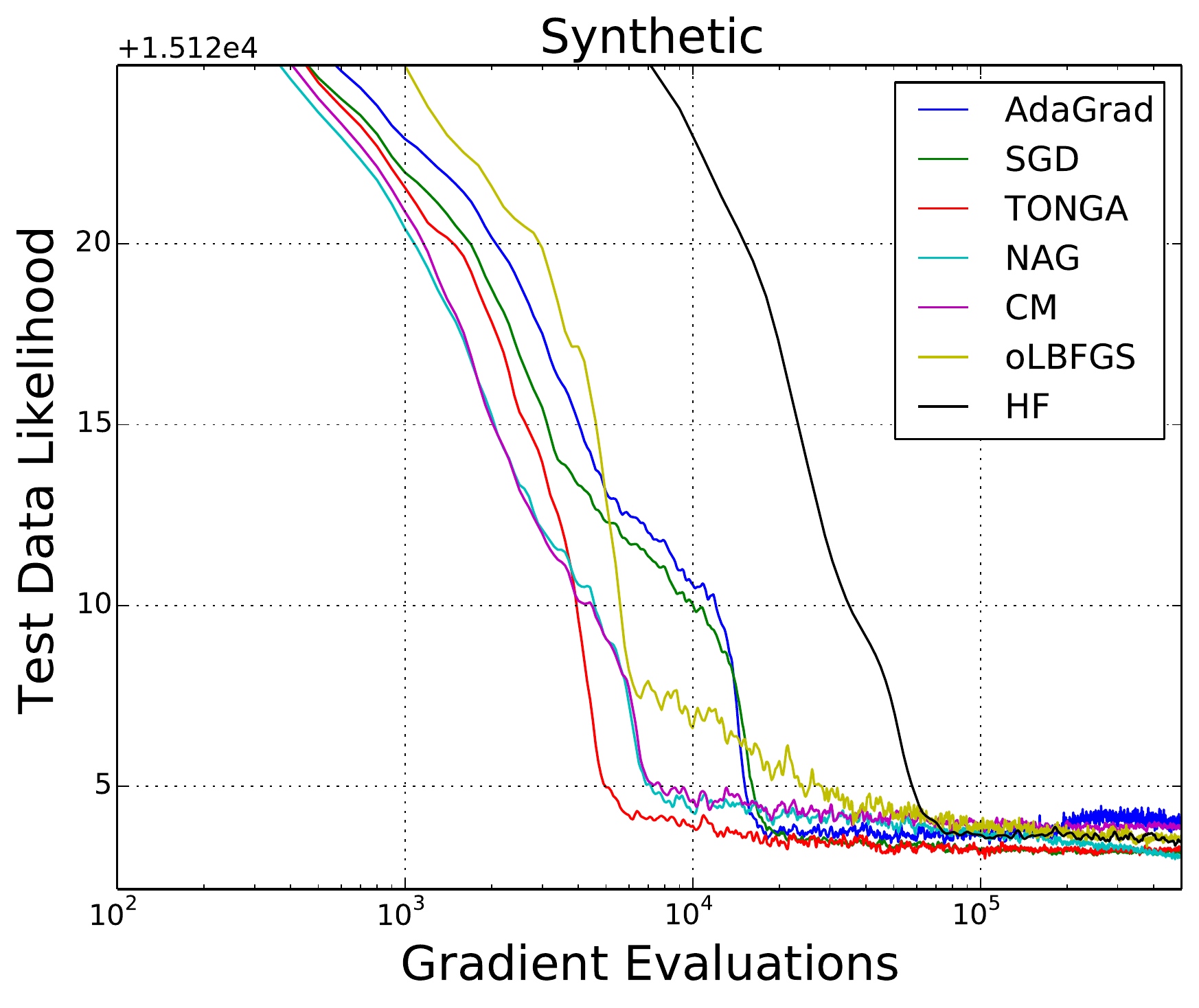} 
\includegraphics[height=0.198\textwidth]{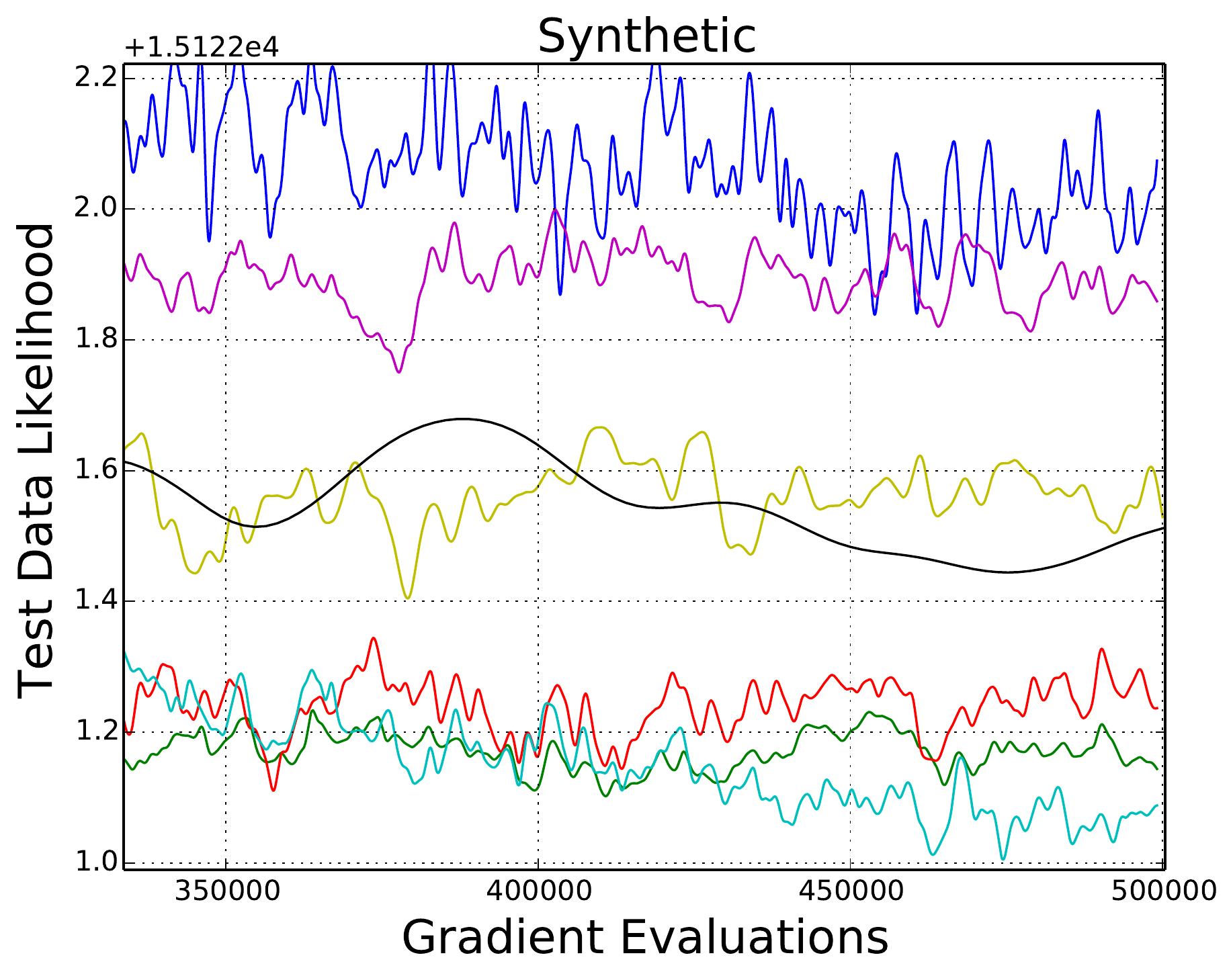} \hfill
\includegraphics[height=0.198\textwidth]{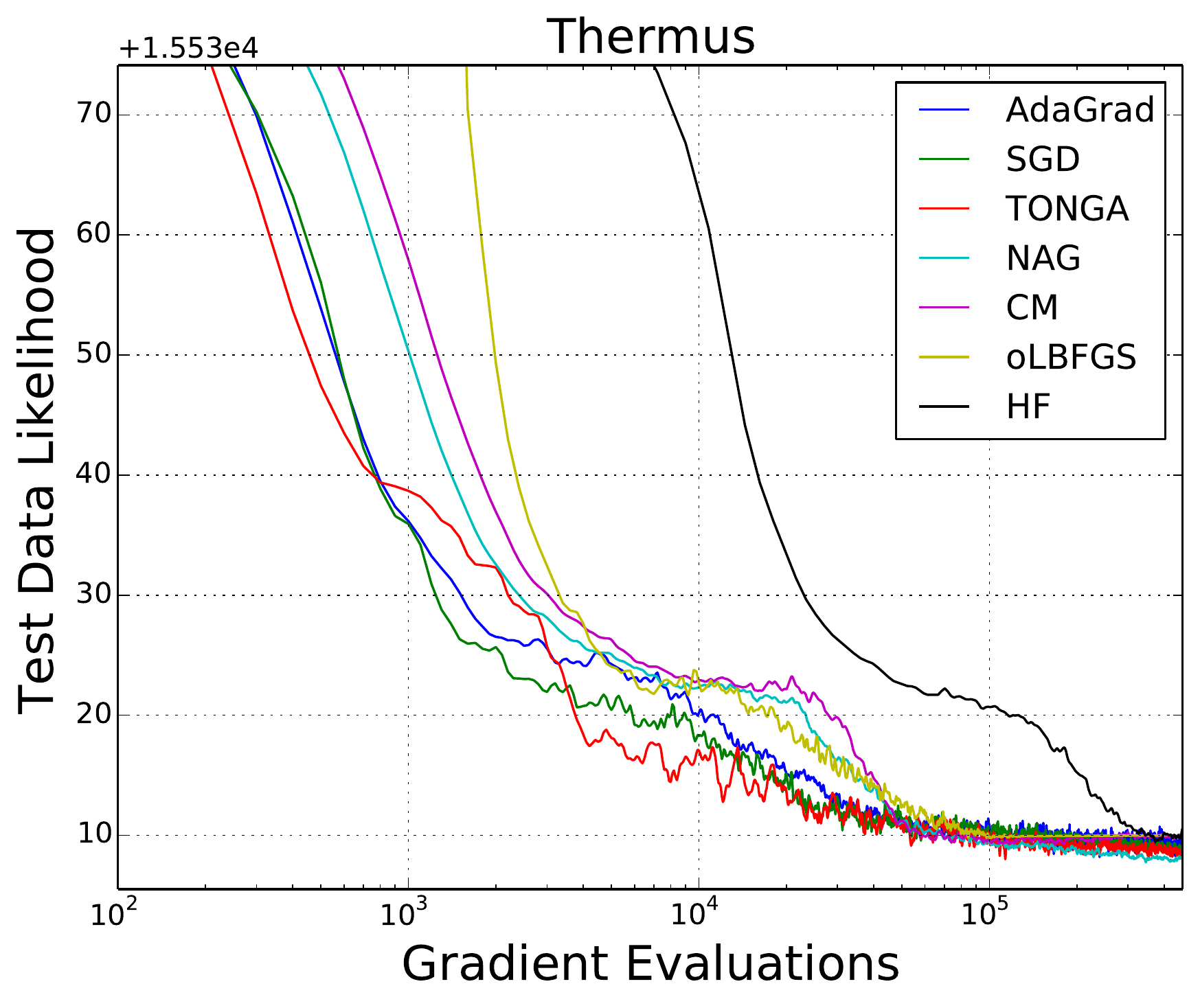} 
\includegraphics[height=0.198\textwidth]{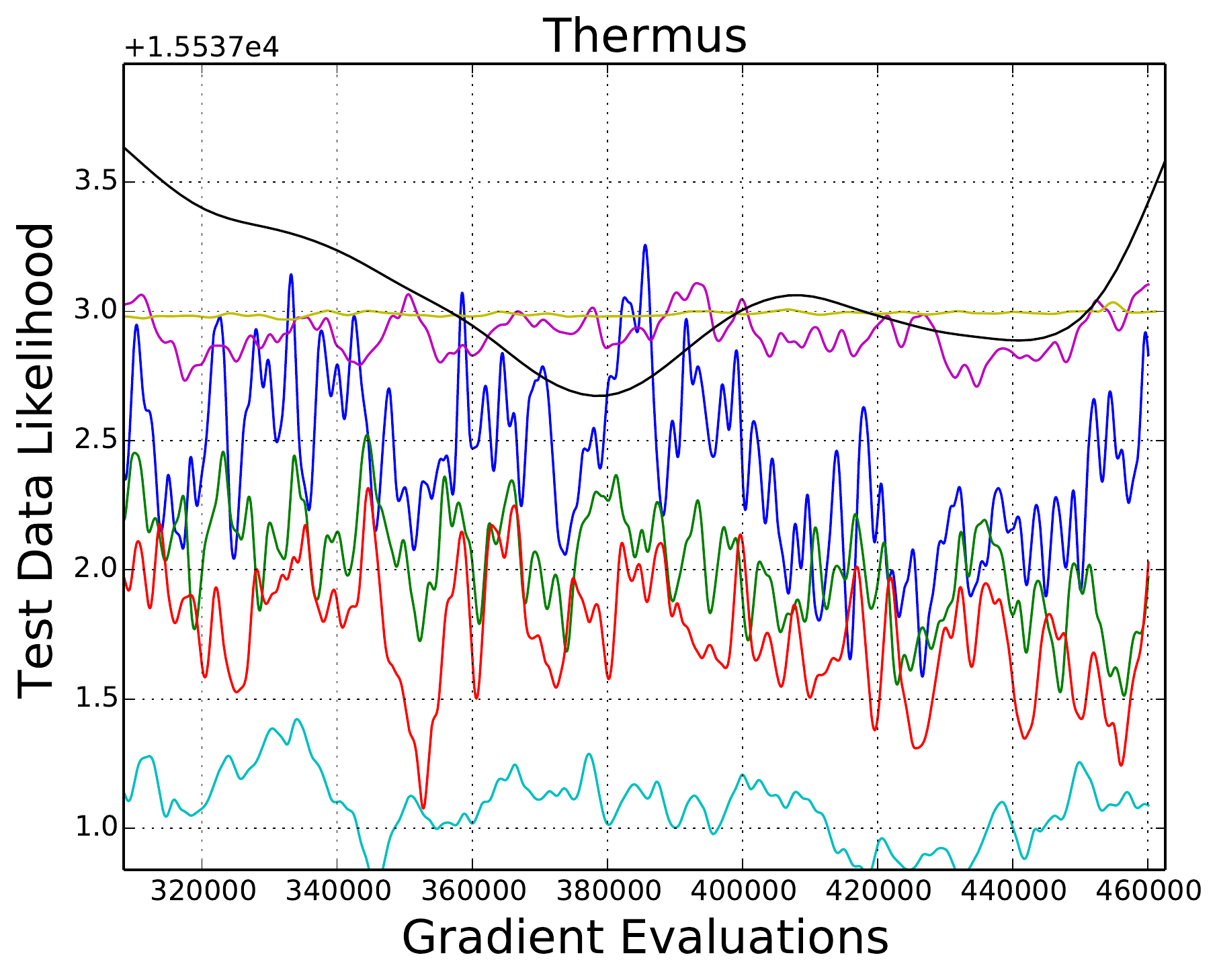}

\caption{\label{fig:resplots}
The negative log likelihood of the test data versus the number of
gradient evaluations for the synthetic (left) and Thermus (right) datasets with
a zoomed in view of the final iterations.
}
\vspace{-0.2cm}
\end{figure}

\subsection{Discussion}
Before presenting our results we discuss some general 
observations about this objective function.
First, in the neighbourhood of a solution $\density^*$ which explains each particle
image well enough that only a single term (corresponding to a single orientation 
and shift) in Equation \ref{eq:ApproxMarginalization} is significant, Equation
\ref{eq:OptimProblem} becomes the sum of the log prior term and $K$ quadratic
terms in $\density$.
Thus, near $\density^*$, the objective is roughly an L1 regularized
linear least squares problem and thus is approximately convex.

In general, the objective function has the algebraic form of a mixture model
posterior due to the sum in Equation \ref{eq:ApproxMarginalization}.
Each orientation is analogous to a mixture component, with parameters corresponding
to the respective slice of $\ftdensity$.
One might suspect the typical pathologies of mixture models to manifest
such as overfitting when some components
(\ie, orientations) are assigned only a small number of data points.
However unlike a mixture model, in this problem the parameters of each component 
are interrelated, as each is a linear combination of elements of $\ftdensity$.
In particular, low frequency coefficients are shared with many mixture components, 
while progressively higher frequencies are shared by fewer and fewer.

Taken together, these observations suggest that, while the objective function
in Equation \ref{eq:OptimProblem} is not convex, it should be 
well behaved so long as the low-frequency Fourier coefficients are
approximately correct.
This observation can also be seen as further motivation for only considering
Fourier coefficients below a threshold, as we expect the restricted problem to
be better behaved.
In practice, when higher resolution structures are sought a good strategy
would be to introduce high-frequency Fourier coefficients gradually.

\section{Experiments}
%
To explore the potential of different optimization methods we acquired two 
datasets, one synthetic and one real.
The real dataset is of ATP synthase from the \emph{Thermus
thermophilus} bacteria, which is a large transmembrane molecule which
provides energy to cells.
The dataset, consisting of $46,105$ particle images and CTF parameters, was
provided by \citet{Lau2012}.
The high resolution structure from \citep{Lau2012} and some sample images 
are shown in Figure \ref{fig:cryoimgs}.
As a specimen for Cryo-EM this dataset is known to be challenging as there 
is no particle symmetry and the contrast is low.
The second dataset was synthetically created using a simple, hand crafted 
electron density.
Using that density, $50,000$ particle images were generated by uniformly
sampling random orientations and assuming zero in-plane translation.
CTFs were simulated with parameters randomly selected from the Thermus dataset
and noise was added to have a comparable SNR.
The synthetic density and some simulated images are shown in Figure
\ref{fig:phantom}.

\newcommand{\incphantom}[2]{\includegraphics[height=0.1125\textwidth]{figs/Phantom_#1_#2.png}}
\newcommand{\incthermus}[2]{\includegraphics[height=0.1125\textwidth]{figs/Thermus_#1_#2.png}}
\newcommand{\alghdr}[1]{\rotatebox{90}{\centering #1}}
\begin{figure}
\centering
\begin{tabular}{c|ccc||ccc|}

& \multicolumn{3}{c|}{Synthetic (50K Images)}
& \multicolumn{3}{|c|}{Thermus (46K Images)} \\
\hline\hline
 & $5$K & $50$K & Final
 & $5$K & $50$K & Final \\
\hline 

\alghdr{\ \ \ \ SGD} &
\incphantom{SGD}{50} &
\incphantom{SGD}{500} &
\incphantom{SGD}{4990} &
\incthermus{SGD}{50} &
\incthermus{SGD}{500} &
\incthermus{SGD}{4600} \\
\hline 

\alghdr{\ AdaGrad} &
\incphantom{AdaGrad}{50} &
\incphantom{AdaGrad}{500} &
\incphantom{AdaGrad}{4990} &
\incthermus{AdaGrad}{50} &
\incthermus{AdaGrad}{500} &
\incthermus{AdaGrad}{4600} \\
\hline 

\alghdr{\ \ \ \ \ CM} &
\incphantom{CM}{50} &
\incphantom{CM}{500} &
\incphantom{CM}{4990} &
\incthermus{CM}{50} &
\incthermus{CM}{500} &
\incthermus{CM}{4600} \\
\hline 

\alghdr{\ \ \ \ NAG} &
\incphantom{NAG}{50} &
\incphantom{NAG}{500} &
\incphantom{NAG}{4990} &
\incthermus{NAG}{50} &
\incthermus{NAG}{500} &
\incthermus{NAG}{4600} \\
\hline 

\alghdr{\ \ TONGA} &
\incphantom{TONGA}{50} &
\incphantom{TONGA}{500} &
\incphantom{TONGA}{4990} &
\incthermus{TONGA}{50} &
\incthermus{TONGA}{500} &
\incthermus{TONGA}{4600} \\
\hline

\alghdr{oLBFGS} & 
\incphantom{oLBFGS}{30} &
\incphantom{oLBFGS}{50} &
\incphantom{oLBFGS}{2490} &
\incthermus{oLBFGS}{30} &
\incthermus{oLBFGS}{50} &
\incthermus{oLBFGS}{2300} \\
\hline

\alghdr{\ \ \ \ \ \ \ HF} &
\incphantom{HF}{50} &
\incphantom{HF}{100} &
\incphantom{HF}{4990} &
\incthermus{HF}{50} &
\incthermus{HF}{100} &
\incthermus{HF}{4600} \\
\hline
\end{tabular}
\caption{\label{fig:resphantom}
Estimated structures after $5$K, $50$K and all gradient evaluations (10
epochs) on two datasets, using seven different stochastic optimization methods.
}
\vspace{-0.2cm}
\end{figure}

\paragraph{Optimization Methods.}
Optimization was performed on these datasets with seven different stochastic 
optimization algorithms.
Specifically, we used traditional stochastic gradient descent (SGD), SGD with
classical momentum (CM) \citep{Polyak1964}, SGD with Nesterovs Accelerated
Gradient (NAG) \citep{Nesterov1983,Sutskever2013}, AdaGrad \citep{Duchi2011},
TONGA \citep{LeRoux2008}, Online LBFGS \citep{Schraudolph2007} and Hessian-Free
optimization (HF) \citep{Martens2010}.
Particle images for both datasets are $128 \times 128$ pixels and their values
were rescaled by the standard deviation of noise so that $\noiseStd=1$.
Prior parameters were set to $\lambda_+ = 10^{-4}$ and $\lambda_- = 10^{-4}$ for 
the synthetic dataset and $\lambda_+ = 3\times10^{-2}$ and $\lambda_- = 10^{-4}$ for
the Thermus dataset.
Fourier coefficients were considered out to a radius of 10\% of the Nyquist limit
for the synthetic dataset and 16\% for Thermus.
A minibatch size of $100$ was used for all methods except HF.
For HF, 5 conjugate gradient iterations were used with a minibatch size of
$300$ and a damping parameter which was tuned by hand.
The larger minibatch size was chosen as a trade-off; smaller minibatches
required strong damping which resulted in small step sizes and slower
convergence.

The base learning rate was tuned by hand for each method by examining
performance on a subset of the training examples.
For all methods except AdaGrad and HF, the learning rate is annealed
according to $\eta_t = \eta_0 ( 1 + \lambda t )^{-0.75}$ where
$\lambda = 10^{-2}$, $t$ is the iteration number and $\eta_0$ is a base
learning rate.
The momentum parameters for CM and NAG were set to
$\min \{0.9, 1 - 2^{-1-\log_2 ( \lfloor \frac{t}{100} \rfloor + 1)}\}$ 
\citep{Sutskever2013}.
For TONGA the covariance matrix was approximated with a rank $20$ approximation
after every $20$ iterations and the regularization parameter was tuned along
with the learning rate.
\begin{wrapfigure}{r}{0.225\textwidth}
\centering
\includegraphics[width=0.2\textwidth]{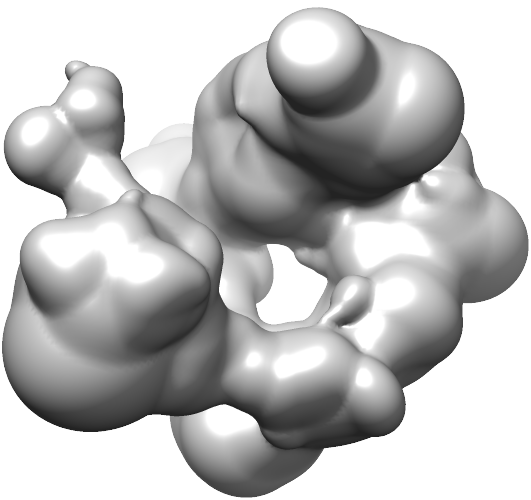}
\caption{\label{fig:initialization}
The random initialization used in all experiments, generated by
summing random spheres.}
\vspace{-0.5cm}
\end{wrapfigure}

Online LBFGS used $m=30$ update vectors to approximate the inverse Hessian.
All methods were run for $10$ epochs except for HF and Online LBFGS which were
run for fewer iterations to account for the increased number of gradient
evaluations needed for these methods.

For both datasets $100$ particle images were randomly held out as a test set
and the negative log likelihood on the test set used to measure performance.
The remaining images were randomly shuffled with each method seeing the images
in the same order.
In each case, optimization was initialized with the same randomly constructed 
density shown in Figure \ref{fig:initialization}.

\paragraph{Stochastic Optimization Results.}
Results of the optimizations are plotted in Figure \ref{fig:resplots} versus
the number of gradients of individual particle images.
This was used to account for the additional gradient evaluations required by
Online LBFGS and the different minibatch sizes and Hessian-vector products
(implemented with a directional finite difference) required by HF.
We show the estimated densities throughout optimization in Figure
\ref{fig:resphantom}.
Note that all algorithms were able to find reasonable solutions
eventually, with the final data likelihood of all methods being within
two log units of each other.
In terms of speed, an epoch (or equivalent) took approximately one hour
using 32 threads on a quad, eight core 2.9GHz  Intel Xeon CPU.
Looking at the plots in Figure \ref{fig:resplots} we can
make some observations about the efficacy of individual methods.

First, note that behaviour is different between the synthetic and
real datasets.
We suspect that outliers in the Thermus dataset are to blame for part of this
difference.
While the image formation model described in the previous section is a good
approximation which has been well established in the Cryo-EM literature, it is
not perfect.
Further, the particle images were manually selected from large micrographs
and mistakes can be made due the high signal-to-noise ratio as well as simple
human error.
An additional difference between the two datasets is the shape of the structure
itself.
The strong cylindrical shape of the synthetic structure causes the objective
function to have directions of low curvature at early iterations resulting
from this symmetry.
Methods which are able to efficiently traverse through these regions are likely
to do better.
Note that structures like this are not uncommon in nature, as many molecules
of interest, in particular viruses, have strong symmetries.

With this in mind, we note that TONGA and the momentum methods
(CM and NAG) do well on the synthetic dataset as might have been predicted
while AdaGrad and SGD clearly slow down around $10^4$ as they traverse this
low-curvature area.
Online LBFGS does not appear to plateau in the same way as SGD and AdaGrad,
but the added cost of the extra gradient evaluations needed to compute the
update vectors for LBFGS are not outweighed by the faster progress it is
able to make at each iteration.
In comparison, on the Thermus dataset this low curvature area does not appear
to be present due to the asymmetry of the particle, resulting in a similar
qualitative behaviour of the methods.
That said, SGD, AdaGrad and TONGA generally converge faster than Online LBFGS
or the momentum methods.
For both datasets, however, the story with Hessian-Free optimization is
consistent: the added cost of the CG iterations and larger minibatches may allow 
faster progress in a given iteration, but that progress does not outweigh the
additional computational cost.

\paragraph{Comparison to State-of-the-Art.}
To compare this method to existing methods for structure determination, 
we selected two approaches.
The first is a standard iterative projection matching scheme where
images are matched to an initial density through a global cross-correlation
search.
The density is then reconstructed based on these orientations and this
process is iterated.
The second is the RELION package described in \citep{Scheres2012} which
uses a similar marginalized model as our method but with a batch
EM algorithm to perform optimization.
We used publicly available code for both of these approaches and 
initialized using the density shown in Figure \ref{fig:initialization}.

\begin{figure}
\centering
\begin{tabular}{ccc}
\includegraphics[height=0.19\textwidth]{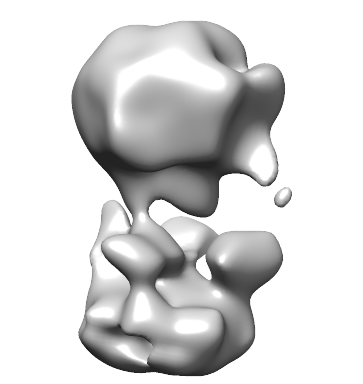} &
\includegraphics[height=0.19\textwidth]{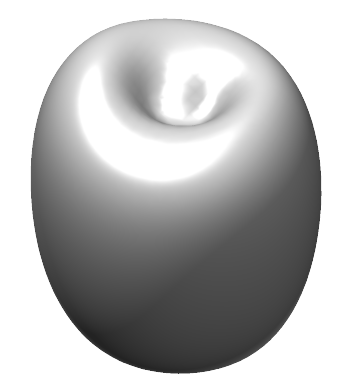} &
\includegraphics[height=0.19\textwidth]{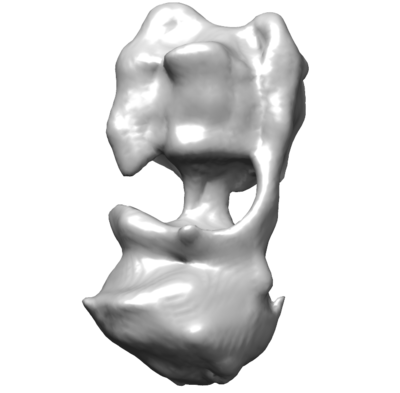} \\
{\tiny Projection Matching} &
{\tiny RELION} &
{\tiny Proposed Approach} \\ 
{\tiny 5 Epochs} &
{\tiny 5 Epochs} &
{\tiny 1 Epoch} 
\end{tabular}
\caption{\label{fig:baselines}
Baseline comparisons to two existing standard methods.
Iterative projection matching and reconstruction (left) and
RELION \citep{Scheres2012} (middle) after 5 batch iterations.
The proposed method (right) with SGD after just one epoch.
}
\vspace{-0.2cm}
\end{figure}

We ran each method for five iterations, roughly equivalent computationally
to five epochs or $250,000$ gradient evaluations for our method.
The results after five iterations of these methods are shown in Figure
\ref{fig:baselines}.
In both cases the approaches failed to converge in the time allotted,
particularly when compared with the structures estimated by the proposed
approach with approximately a fifth of computation, \ie, the $50,000$ columns in
Figure \ref{fig:resphantom}.

\section{Conclusions}
This paper has introduced and motivated the challenging problem of 
density estimation for Cryo-EM, formulated it using a probabilistic image
formation model and cast the resulting MAP estimation problem 
as a stochastic optimization.
We have implemented an array of seven stochastic optimization methods ranging
from simple SGD through momentum methods to more complex quasi-Newton methods
and compared their behaviour and performance on the CryoEM problem with both
real and synthetic data.
A comparison of our proposed stochastic optimization approach with current 
methods demonstrates that stochastic optimization yields significant speedups
and robustness to initialization.
The proposed approach converges to reasonable structures in as
little as one epoch from a random initialization while, in comparison,
existing methods make slow progress even after many epochs, or become trapped
in local minima.
Among the stochastic optimization methods tested, we find that all
are effective at finding a good solution, though some typically converge
faster than others. 
Notably we find that methods which require significant additional computation per
iteration like Online LBFGS \citep{Schraudolph2007} or Hessian-free optimization  
\citep{Martens2010} do not perform better than simply taking more iterations
of other methods on this problem.

We believe that the problem of density estimation for CryoEM is an important
problem and an interesting challenge for stochastic optimization algorithms in
general.
In particular, this work highlights one of the major limitations of 
most existing methods: their reliance on manual tuning.
All methods had parameters which required some amount of manual tuning and
every dataset would potentially require those parameters to be re-adjusted.
While efforts were made to standardize the datasets, some amount of 
retuning is generally necessary for each new dataset.
More research is needed to develop methods which are able to automatically
adapt to different objective functions and we note that there are 
several methods which show significant promise along these lines
\citep{Duchi2011,LeRoux2012,Schmidt2013a,Schaul2013}.

\bibliography{refs}
\bibliographystyle{plainnat}

\end{document}